\documentclass[10pt,twocolumn,letterpaper]{article}
\pdfoutput=1

\usepackage[utf8]{inputenc}
\usepackage[T1]{fontenc}

\usepackage{cvpr}
\usepackage{times}
\usepackage{graphicx}

\usepackage{csquotes}
\usepackage{booktabs}
\usepackage{microtype}
\usepackage{tikz}
\usepackage{nicefrac}
\usepackage{mathtools}
\usepackage{todonotes}

\usetikzlibrary{arrows}
\usetikzlibrary{calc}
\usetikzlibrary{decorations.markings}
\usetikzlibrary{decorations.text}

\usepackage[pagebackref=true,breaklinks=true,letterpaper=true,colorlinks,bookmarks=false]{hyperref}

\cvprfinalcopy % *** Uncomment this line for the final submission

\def\cvprPaperID{2230} % *** Enter the CVPR Paper ID here

\ifcvprfinal\fi

\def\citet{\cite}
\def\citep{\cite}

\begin{document}

%%%%%%%%% TITLE
\title{Full Resolution Image Compression with Recurrent Neural Networks}

\author{George Toderici\\
Google Research\\
{\tt\small gtoderici@google.com}
\and
Damien Vincent\\
{\tt\small damienv@google.com}
\and
Nick Johnston\\
{\tt\small nickj@google.com}
\and
Sung Jin Hwang\\
{\tt\small sjhwang@google.com}
\and
David Minnen\\
{\tt\small dminnen@google.com}
\and
Joel Shor\\
{\tt\small joelshor@google.com}
\and
Michele Covell\\
{\tt\small covell@google.com}
}

\maketitle
%\thispagestyle{empty}

%%%%%%%%% ABSTRACT
\begin{abstract}
This paper presents a set of full-resolution lossy image
compression methods based on neural networks. Each of the architectures we
describe can provide variable compression rates during deployment without
requiring retraining of the network: each network need only be trained once.
All of our architectures consist of a recurrent neural network (RNN)-based
encoder and decoder, a binarizer, and a neural network for entropy coding.
We compare RNN types (LSTM, associative LSTM) and introduce a new hybrid of GRU and ResNet.
We also study ``one-shot'' versus additive reconstruction architectures and introduce a new scaled-additive framework.
We compare to
previous work, showing improvements of 4.3\%--8.8\% AUC (area under the rate-distortion curve), depending on the perceptual
metric used.
As far as we know,
this is the first neural network architecture that is able to outperform
JPEG at image compression across most bitrates on the rate-distortion curve on the Kodak dataset
images, with and without the aid of entropy coding.
\end{abstract}

%%%%%%%%% BODY TEXT
\section{Introduction}
Image compression has traditionally been one of
the tasks which neural networks were suspected to be good at, but
there was little evidence that it would be possible to train a single
neural network that would be competitive across compression rates
and image sizes.
\citet{toderici2015variable} showed that it is possible
to train a single recurrent neural network and achieve better
than state of the art compression rates for a given quality regardless of
the input image, but was
limited to 32$\times$32 images. In that work, no effort was made to capture
the long-range dependencies between image patches.

Our goal is to provide a neural network which is competitive across compression
rates on images of arbitrary size. There are two possible ways to achieve this:
1) design a stronger patch-based residual encoder; and 2) design
an entropy coder that is able to capture long-term dependencies between patches
in the image. In this paper, we address both problems and combine
the two possible ways to improve compression rates for a given quality.

In order to measure how well our architectures are doing (i.e., ``quality''), we cannot rely
on typical metrics such as Peak Signal to Noise Ratio (PSNR), or $L_p$ differences
between compressed and reference images because the human visual system is
more sensitive to certain types of distortions than others. This idea was exploited
in lossy image compression methods such as JPEG. In order to be able to measure
such differences, we need to use a human visual system-inspired measure which,
ideally should correlate with how humans perceive image differences. Moreover,
if such a metric existed, and were differentiable, we could directly optimize
for it. Unfortunately, in the literature there is a wide variety of metrics of varying quality,
most of which are non-differentiable. For evaluation purposes, we
selected two commonly used metrics, PSNR-HVS \citep{psnrhvs} and MS-SSIM \citep{wang2003multiscale}, as discussed in~\autoref{section:results}.

\subsection{Previous Work}
Autoencoders have been used to reduce the dimensionality of images
\citep{Hinton2006},
convert images to compressed binary codes for retrieval
\citep{Krizhevsky2011}, and to extract compact visual representations that
can be used in other applications \citep{vincent2010}. More recently,
 variational (recurrent) autoencoders have been directly applied to the problem of
compression \citep{gregor2016conceptual} (with results on images of size up to
64$\times$64 pixels), while non-variational recurrent neural networks were used
to implement variable-rate encoding \citep{toderici2015variable}.

Most image compression neural networks use a fixed compression rate based on the
size of a bottleneck layer \citep{balle2016}.
This work extends previous methods by supporting variable rate compression while maintaining high compression rates beyond thumbnail-sized images.

\section{Methods}
In this section, we describe the high-level model architectures we explored.
The subsections provide additional details about the different recurrent network components in our experiments.
Our compression networks are comprised of an encoding network $E$, a binarizer $B$ and a decoding network $D$,
where $D$ and $E$ contain recurrent network components.
The input images are first encoded, and then transformed into binary codes that can be stored or transmitted to the decoder.
The decoder network creates an estimate of the original input image based on the received binary code.
We repeat this procedure with the residual error, the difference between the original image and the reconstruction from the decoder.

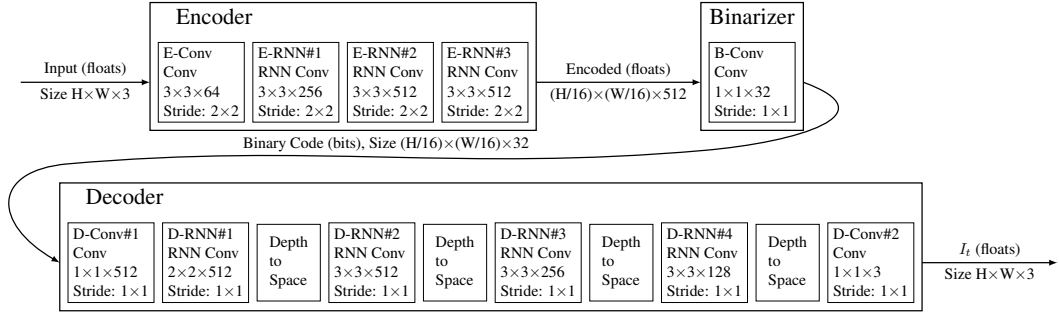
\begin{figure*}
  \centering
  \scalebox{.6}{%
  \begin{tikzpicture}[every text node part/.style={align=left},block/.style={draw,rectangle,minimum height=1.7cm,minimum width=1.4cm}]
    \node (E-conv) at (0,0) [block] {E-Conv\\Conv\\3$\times$3$\times$64\\Stride: 2$\times$2};
    \node (E-rnn1) at ([xshift=5pt]E-conv.east) [anchor=west,block] {E-RNN\#1\\RNN Conv\\3$\times$3$\times$256\\Stride: 2$\times$2};
    \node (E-rnn2) at ([xshift=5pt]E-rnn1.east) [anchor=west,block] {E-RNN\#2\\RNN Conv\\3$\times$3$\times$512\\Stride: 2$\times$2};
    \node (E-rnn3) at ([xshift=5pt]E-rnn2.east) [anchor=west,block] {E-RNN\#3\\RNN Conv\\3$\times$3$\times$512\\Stride: 2$\times$2};
    \node (E) at ([shift=({8pt,8pt})]E-conv.north west) [anchor=south west] {\Large Encoder};
    \coordinate (Ebox-left) at ([shift=({-5pt,-5pt})]E-conv.south west);
    \draw [thick] (Ebox-left) rectangle ([xshift=5pt]E-rnn3.east |- E.north) coordinate (Ebox-right);

    \draw [thick,-latex] (-4,0 |- E-conv) -- (Ebox-left |- E-conv) node [midway,above] {Input (floats)} node [midway,below] {Size H$\times$W$\times$3};

    \node (B-conv) at ([xshift=4cm]E-rnn3.east) [anchor=west,block] {B-Conv\\Conv\\1$\times$1$\times$32\\Stride: 1$\times$1};
    \node (B) at ([yshift=8pt]B-conv.north) [anchor=south] {\Large Binarizer};
    \coordinate (Bbox-left) at ([shift=({-5pt,-5pt})]B-conv.south west);
    \draw [thick] (Bbox-left) rectangle ([xshift=5pt]B-conv.east |- E.north) coordinate (Bbox-right);

    \draw [thick,-latex] (Ebox-right |- E-rnn3) -- (Bbox-left |- B-conv) node [midway,above] {Encoded (floats)} node [midway,below] {(H/16)$\times$(W/16)$\times$512};

    \node (D-conv1) at (-2,-4) [block] {D-Conv\#1\\Conv\\1$\times$1$\times$512\\Stride: 1$\times$1};
    \node (D-rnn1) at ([xshift=5pt]D-conv1.east) [anchor=west,block] {D-RNN\#1\\RNN Conv\\2$\times$2$\times$512\\Stride: 1$\times$1};
    \node (D-d2s1) at ([xshift=5pt]D-rnn1.east) [anchor=west,block] {Depth\\to\\Space};
    \node (D-rnn2) at ([xshift=5pt]D-d2s1.east) [anchor=west,block] {D-RNN\#2\\RNN Conv\\3$\times$3$\times$512\\Stride: 1$\times$1};
    \node (D-d2s2) at ([xshift=5pt]D-rnn2.east) [anchor=west,block] {Depth\\to\\Space};
    \node (D-rnn3) at ([xshift=5pt]D-d2s2.east) [anchor=west,block] {D-RNN\#3\\RNN Conv\\3$\times$3$\times$256\\Stride: 1$\times$1};
    \node (D-d2s3) at ([xshift=5pt]D-rnn3.east) [anchor=west,block] {Depth\\to\\Space};
    \node (D-rnn4) at ([xshift=5pt]D-d2s3.east) [anchor=west,block] {D-RNN\#4\\RNN Conv\\3$\times$3$\times$128\\Stride: 1$\times$1};
    \node (D-d2s4) at ([xshift=5pt]D-rnn4.east) [anchor=west,block] {Depth\\to\\Space};
    \node (D-conv2) at ([xshift=5pt]D-d2s4.east) [anchor=west,block] {D-Conv\#2\\Conv\\1$\times$1$\times$3\\Stride: 1$\times$1};
    \node (D) at ([shift=({8pt,8pt})]D-conv1.north west) [anchor=south west] {\Large Decoder};
    \coordinate (Dbox-left) at ([shift=({-5pt,-5pt})]D-conv1.south west);
    \draw [thick] (Dbox-left) rectangle ([xshift=5pt]D-conv2.east |- D.north) coordinate (Dbox-right);

    % \draw [thick,-latex] (Bbox-right |- B-conv) to[out=-20,in=5,looseness=2] ([yshift=1.6cm]D-rnn4.north) [postaction={decorate,decoration={text along path,text align=center,text={Binary code}}}] to[out=-175,in=5] ([yshift=1.3cm]D-conv1.north west) to[out=-175,in=150,looseness=2] (Dbox-left |- D-conv1);
    \draw [thick,-latex] (Bbox-right |- B-conv) to[out=-20,in=5,looseness=2] ([yshift=1.6cm]D-rnn4.north) to[out=-175,in=5] node [above] {Binary Code (bits), Size (H/16)$\times$(W/16)$\times$32} ([yshift=1.3cm]D-conv1.north west) to[out=-175,in=150,looseness=2] (Dbox-left |- D-conv1);

    \draw [thick,-latex] (Dbox-right |- D-conv2) -- ++(3,0) node [midway,above] {$I_t$ (floats)} node [midway,below] {Size H$\times$W$\times$3};
    \useasboundingbox ([shift={(-0.5cm,-0.5cm)}]current bounding box.south west) -- (current bounding box.north east);
    \useasboundingbox ([shift={(0.5cm,0.5cm)}]current bounding box.north east) -- (current bounding box.south west);
  \end{tikzpicture}}
  \caption{A single iteration of our shared RNN architecture.}
  \label{fig:general-loop}
\end{figure*}

\autoref{fig:general-loop} shows the architecture of a single iteration of our model.
While the network weights are shared between iterations, the states in the recurrent components are propagated to the next iteration.
Therefore residuals are encoded and decoded in different contexts in different iterations.
Note that the binarizer $B$ is stateless in our system.

We can compactly represent a single iteration of our networks as follows:
\begin{align}\label{eq:eq0}
  b_t = B(E_t(r_{t-1})),&
  \quad
  \hat x_t = D_t(b_t) + \gamma \hat x_{t-1},
  \quad \\
  r_t = x - \hat x_t,&
  \quad
  r_0 = x,
  \quad
  \hat x_0 = 0
\end{align}
where $D_t$ and $E_t$ represent the decoder and encoder with their states at iteration $t$ respectively,
$b_t$ is the progressive binary representation; $\hat x_t$ is the progressive reconstruction of the original image $x$ with $\gamma = 0$ for \enquote{one-shot} reconstruction or 1 for additive reconstruction (see \autoref{sec:recon frame});
and $r_t$ is the residual between $x$ and the reconstruction $\hat x_t$.
In every iteration, $B$ will produce a binarized bit stream $b_t \in \{-1, 1\}^m$ where $m$ is the number of bits produced after every iteration, using the approach reported in \cite{toderici2015variable}.
After $k$ iterations, the network produces $m \cdot k$ bits in total.
Since our models are fully convolutional, $m$ is a linear function of input size.  For image patches of 32$\times$32, $m=128$.

The recurrent units used to create the encoder and decoder include two convolutional kernels: one on the input vector which comes into
the unit from the previous layer and the other one on the state vector which provides the recurrent nature of the unit.  We will refer to the
convolution on the state vector and its kernel as the ``hidden convolution'' and the ``hidden kernel''.

In \autoref{fig:general-loop},
we give the spatial extent of the input-vector convolutional kernel along with the output depth.  All convolutional kernels allow full mixing across depth.
For example, the unit {\it D-RNN\#3} has 256 convolutional kernels that operate on the input vector, each with 3$\times$3 spatial extent and full input-depth
extent (128 in this example, since the depth of {\it D-RNN\#2} is reduced by a factor of four as it goes through the ``Depth-to-Space'' unit).

The spatial extents of the hidden kernels are all 1$\times$1, except for in units {\it D-RNN\#3} and {\it D-RNN\#4} where the hidden kernels are 3$\times$3.
The larger hidden kernels consistently resulted in improved compression curves compared to the 1$\times$1 hidden kernels exclusively used in \cite{toderici2015variable}.

During training, a $L_1$ loss is calculated on the weighted residuals generated at each iteration (see \autoref{section:results}), so our total loss for the network is:
\begin{equation}
  \beta \sum_t \left\lvert r_t\right\rvert
\label{eq:loss}
\end{equation}

In our networks, each 32$\times$32$\times$3 input image is reduced to a 2$\times$2$\times$32 binarized representation per iteration.
This results in each iteration representing $\nicefrac18$ bit per pixel (bpp).  If only the first iteration is used, this would be 192:1 compression,
even before entropy coding (\autoref{section:entropy}).

We explore a combination of recurrent unit variants and reconstruction frameworks for our compression systems.
We compare these compression results to the results from the deconvolutional network described in \cite{toderici2015variable}, referred to in this paper as the Baseline network.

\vspace{-0.3em}
\subsection{Types of Recurrent Units}
\vspace{-0.3em}

In this subsection, we introduce the different types of recurrent units that we examined.

\textbf{LSTM:} One recurrent neural-network element we examine is a LSTM \citep{lstm:1997} with the formulation proposed by \citet{zaremba2014recurrent}.
Let $x_t$, $c_t$, and $h_t$ denote the input, cell, and hidden states, respectively, at iteration $t$.
Given the current input $x_t$, previous cell state $c_{t-1}$, and previous hidden state $h_{t-1}$, the new cell state $c_t$ and the new hidden state $h_t$ are computed as
\begin{align}
  [f,i,o,j]^T
  &=
  [\sigma,\sigma,\sigma,\tanh]^T
  \big(( W x_t + U h_{t-1} ) + b\big),
  \\
  c_t &= f \odot c_{t-1} + i \odot j,
  \\
  h_t &= o \odot \tanh(c_t),
\end{align}
where $\odot$ denotes element-wise multiplication, and $b$ is the bias.
The activation function $\sigma$ is the sigmoid function $\sigma(x) = 1 / (1 + \exp(-x))$.
The output of an LSTM layer at iteration $t$ is $h_t$.

The transforms $W$ and $U$, applied to $x_t$ and $h_{t-1}$, respectively, are convolutional linear transformations.
That is, they are composites of Toeplitz matrices with padding and stride transformations.
The spatial extent and depth of the $W$ convolutions are as shown in \autoref{fig:general-loop}.
As pointed out earlier in this section, the $U$ convolutions have the same depths as the $W$ convolutions. For a more in-depth explanation, see~\cite{toderici2015variable}.

\textbf{Associative LSTM:} Another neural network element we examine is the Associative LSTM \citep{danihelka2016assoc}.
Associative LSTM extends LSTM using holographic representation.
Its new states are computed as
\begin{gather}
  \begin{align}
      [f, &i, o, j, r_i, r_o]^T= \nonumber
      \\
      &[\sigma,\sigma,\sigma,\operatorname{bnd},\operatorname{bnd},\operatorname{bnd}]^T
      \big(( W x_t + U h_{t-1} ) + b\big),
  \end{align}\\
  \begin{align}
    c_t &= f \odot c_{t-1} + r_i \odot i \odot j,
    \\
    h_t &= o \odot \operatorname{bnd}(r_o \odot c_t),
    \\
    \tilde h_t &= ( \text{Re}\, h_t, \text{Im}\, h_t ).
  \end{align}
\end{gather}
The output of an Associative LSTM at iteration $t$ is $\tilde h_t$.
The input $x_t$, the output $\tilde h_t$, and the gate values $f, i, o$ are real-valued, but the rest of the quantities are complex-valued.
The function $\operatorname{bnd}(z)$ for complex $z$ is $z$ if $\lvert z \rvert \leq 1$ and is $z / \lvert z \rvert$ otherwise.
As in the case of non-associative LSTM, we use convolutional linear transformations $W$ and $U$.

Experimentally, we determined that Associative LSTMs were effective only when used in the decoder.
Thus, in all our experiments with Associative LSTMs, non-associative LSTMs were used in the encoder.

\textbf{Gated Recurrent Units:} The last recurrent element we investigate is the Gated Recurrent Unit~\citep{chung2014empirical} (GRU).
The formulation for GRU, which has an input $x_t$ and a hidden state/output $h_t$, is:
\begin{align}
z_{t} & = \sigma(W_{z} x_t + U_{z} h_{t-1}),\\
r_{t} & = \sigma(W_{r} x_t + U_{r} h_{t-1}),\\
h_{t} & = (1 - z_{t}) \odot h_{t-1} + \nonumber\\
      & z_t \odot \tanh(Wx_{t} + U(r_{t} \odot h_{t-1})).
\end{align}

As in the case of LSTM, we use convolutions instead of simple multiplications. Inspired by the core ideas from ResNet~\citep{resnet} and Highway Networks~\citep{srivastava2015highway}, we can think of GRU as a computation block and pass residual information around the block in order to speed up convergence.
Since GRU can be seen as a doubly indexed block, with one index being iteration and the other being space, we can formulate a residual version of GRU which now has two residual connections.
In the equations below, we use $h^{o}_t$ to denote the output of our formulation, which will be distinct from the hidden state $h_t$:
\begin{align}
  h_{t} & = (1 - z_{t}) \odot h_{t-1} + \nonumber \\
        &  z_t \odot \tanh(Wx_{t} + U(r_{t} \odot h_{t-1})) + \alpha_{h} W_h h_{t-1}, \\
  h^{o}_t & = h_t + \alpha_{x} W_{ox} x_t.
\end{align}
where we use $\alpha_{x} = \alpha_{h} = 0.1$ for all the experiments in this paper.

This idea parallels the work done in  Higher Order RNNs \citep{soltani2016higher}, where linear connections are added between iterations, but not between the input and the output of the RNN.

\vspace{-0.5em}
\subsection{Reconstruction Framework}
\vspace{-0.5em}
\label{sec:recon frame}

In addition to using different types of recurrent units, we examine three different approaches
to creating the final image reconstruction from our decoder outputs.
We describe those approaches in this subsection, along with the changes needed to the loss function.

\textbf{One-shot Reconstruction:}
As was done in \cite{toderici2015variable}, we predict the full image after each iteration of the decoder ($\gamma = 0$ in \eqref{eq:eq0}).
Each successive iteration has access to more bits generated by the encoder which allows for a better reconstruction.
We call this approach \enquote{one-shot reconstruction}.
Despite trying to reconstruct the original image at each iteration, we only pass the previous iteration's residual to the next iteration.
This reduces the number of weights, and experiments show that passing both the original image and the residual does not improve the reconstructions.

\textbf{Additive Reconstruction:}
In additive reconstruction, which is more widely used in traditional image coding, each iteration only tries to
reconstruct the residual from the previous iterations.
The final image reconstruction is then the sum of the outputs of all iterations ($\gamma = 1$ in \eqref{eq:eq0}).

\textbf{Residual Scaling:}
In both additive and \enquote{one shot} reconstruction, the residual starts large, and we expect it to decrease with each iteration.
However, it may be difficult for the encoder and the decoder to operate efficiently across a wide range of values.
Furthermore, the rate at which the residual shrinks is content dependent.
In some patches (e.g., uniform regions), the drop-off will be much more dramatic than in other patches (e.g., highly textured patches).

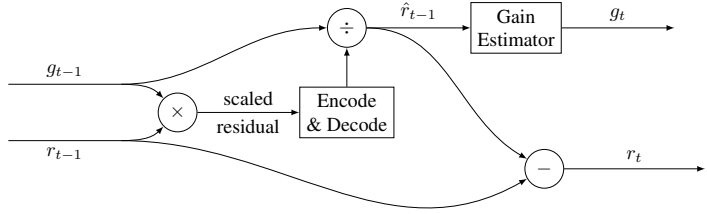
\begin{figure*}
  \centering
  \scalebox{0.75}{%
  \begin{tikzpicture}[every text node part/.style={align=center}]
    \draw (0,0) -- (2,0) node [midway,below,sloped] {$r_{t-1}$};
    \draw (0,1) -- (2,1) node [midway,above,sloped] (B) {$g_{t-1}$};
    \node (C) at (3,0.5) [shape=circle,draw] {$\times$};
    \draw [-latex] (2,0) to[in=-135,out=0] (C);
    \draw [-latex] (2,1) to[in=135,out=0] (C);

    \node (D) at (6,0.5) [shape=rectangle,draw] {Encode\\\& Decode};
    \draw [-latex] (C.east) -- (D.west) node [midway,above] {scaled} node [midway,below] {residual};
    \node (E) at (6,2) [shape=circle,draw] {$\div$};
    \draw [-latex] (D.north) -- (E.south);
    \draw [-latex] (2,1) to[in=180,out=0] (E);

    \node (F) at (9,2) [shape=rectangle,draw] {Gain\\Estimator};
    \draw [-latex] (E.east) -- (F.west) node [midway,above] {$\hat r_{t-1}$};
    \draw [-latex] (F.east) -- ++(2,0) node [pos=0.5,above] {$g_t$};

    \node (G) at (9.5,-0.5) [shape=circle,draw] {$-$};
    \draw [-latex] (E) to[in=150,out=0] (G);

    \draw [-latex] (G.east) -- ++(2.5,0) node [pos=0.5,above] {$r_t$};

    \draw [-latex] (2,0) to[in=-150,out=0] (G);
  \end{tikzpicture}}
  \caption{Adding content-dependent, iteration-dependent residual scaling to the additive reconstruction framework.
  Residual images are of size H$\times$W$\times$3 with three color channels, while gains are of size 1 and the same gain factor is applied to all three channels per pixel.}
  \label{fig:gain}
\end{figure*}

To accommodate these variations, we extend our additive reconstruction architecture to
include a content-dependent, iteration-dependent gain factor.
\autoref{fig:gain} shows the extension that we used.
Conceptually, we look at the reconstruction of the previous residual
image, $r_{t-1}$, and derive a gain multiplier for each patch.
We then
multiply the target residual going into the current iteration by the gain
that is given from processing the previous iteration's output.
\autoref{eq:eq0} becomes:
\begin{gather}
  g_t = G(\hat x_t),
  \quad
  b_t = B(E_t(r_{t-1} \odot \operatorname{ZOH}(g_{t-1}))),
  \quad \\
  \hat r_{t-1} = D_t(b_t) \oslash \operatorname{ZOH}(g_{t-1}),
  \\
  \hat x_t = \hat x_{t-1} + \hat r_{t-1},
  \quad
  r_t = x - \hat x_t, \\
  \quad
  g_0 = 1,
  \quad
  r_0 = x.
\end{gather}
where $\oslash$ is element-wise division and $\operatorname{ZOH}$ is spatial upsampling by zero-order hold.
$G(\cdot)$ estimates the gain
factor, $g_t$, using a five-layer feed-forward convolutional network,
each layer with a stride of two.  The first four layers give an output depth
of 32, using a 3$\times$3 convolutional kernel with an ELU nonlinearity \citep{elu}.
The final layer gives an
output depth of 1, using a 2$\times$2 convolutional kernel, with an ELU nonlinearity.
Since ELU has a range of $(-1,\infty)$ a constant of 2 is added to the output of this network to obtain $g_t$ in the range of $(1,\infty)$.

\vspace{-0.5em}
\section{Entropy Coding}
\label{section:entropy}
\vspace{-0.5em}

The entropy of the codes generated during inference are not maximal because
the network is not explicitly designed to maximize entropy in its codes, and the
model does not necessarily exploit visual redundancy over a large spatial extent.
Adding an entropy coding layer can further improve the compression ratio,
as is commonly done in standard image compression codecs.
In this section, the image encoder is a given and is only used as a binary code generator.
%In this section, the image encoder is assumed to be fully learned and is only used as a binary code generator.

The lossless entropy coding schemes considered here are fully convolutional,
process binary codes in progressive order and for a given encoding iteration in raster-scan order.
All of our image encoder architectures generate binary codes of the form $c(y, x, d)$ of size $H \times W \times D$, where $H$ and $W$
are integer fractions of the image height and width and $D$ is $m$ $\times$ the number of iterations.
We consider a standard lossless encoding framework that combines a conditional probabilistic model of the current binary code
$c(y, x, d)$ with an arithmetic coder to do the actual compression.
More formally, given a context $T(y,x,d)$ which depends only on previous bits
in stream order, we will estimate $P(c(y,x,d) \mid T(y,x,d))$ so that the
expected ideal encoded length of $c(y,x,d)$ is the cross entropy between
$P(c\mid T)$ and $\hat P(c\mid T)$.
We do not consider the small penalty involved by using a practical arithmetic coder that requires a quantized
version of $\hat{P}(c\mid T)$.

\subsection{Single Iteration Entropy Coder}

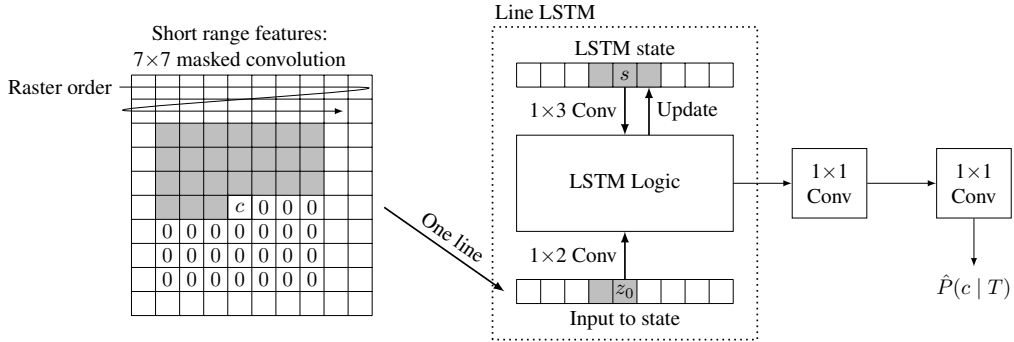
\begin{figure*}
  \centering
  \scalebox{.8}{%
  \begin{tikzpicture}[scale=0.4,every text node part/.style={align=center}]
    \fill [gray!50] (2,5) rectangle ++(7,3) (2,4) rectangle ++(3,1);
    \draw [step=1] (1,0) grid ++(10,10);

    \path (1,10) -- +(9,0) node [midway,above] {Short range features:\\7$\times$7 masked convolution};
    \draw (5,4) rectangle +(1,1) node [pos=.5] {$c$};
    \foreach \x in {6,7,8}
      \path (\x,4) rectangle +(1,1) node [pos=.5] {$0$};
    \foreach \x in {2,...,8}
    \foreach \y in {1,2,3}
      \path (\x,\y) rectangle +(1,1) node [pos=.5] {$0$};

    \draw (0.5,9.5) -- (10,9.5) to[in=180,out=0,looseness=1.5] (1.5,8.5) [-latex] -- (9.8,8.5);
    \node at (0.5,9.5) [left] {Raster order};

    \draw [thick,-latex] (11.5,4.5) -- (16.5,1) node [midway,above,sloped] {One line};

    \begin{scope}[shift={(17,0.5)}]
      \fill [gray!50] (3,0) rectangle +(2,1);
      \draw [step=1] (0,0) grid +(9,1);
      \path (4,0) rectangle +(1,1) node [pos=.5] {$z_0$};
      \path (0,0) -- +(9,0) node [midway,below] {Input to state};

      \draw (0,3) rectangle ++(9,4) node [pos=.5] {LSTM Logic};

      \fill [gray!50] (3,9) rectangle ++(3,1);
      \draw [step=1] (0,9) grid ++(9,1);
      \path (4,9) rectangle ++(1,1) node [pos=.5] {$s$};
      \path (0,10) -- +(9,0) node [midway,above] {LSTM state};

      \draw [thick,-latex] (4.5,1) -- +(0,2) node [midway,left] {1$\times$2 Conv};
      \draw [thick,-latex] (4.5,9) -- +(0,-2) node [midway,left] {1$\times$3 Conv};
      \draw [thick,-latex] (5.5,7) -- +(0,2) node [midway,right] {Update};

      \draw [thick,dotted] (-1,-1.5) rectangle ++(11,13);
      \path (-1,11.5) -- +(11,0) node [pos=.2,above] {Line LSTM};

      \node (F) at (13,5) [draw,shape=rectangle,inner sep=7] {1$\times$1\\Conv};
      \node (G) at (19,5) [draw,shape=rectangle,inner sep=7] {1$\times$1\\Conv};

      \draw [-latex] (9,5) -- (F.west);
      \draw [-latex] (F.east) -- (G.west);
      \draw [-latex] (G.south) -- +(0,-2) node [below] {$\hat P(c\mid T)$};
    \end{scope}
  \end{tikzpicture}}
  \caption{Binary recurrent network (BinaryRNN) architecture for a single iteration. The gray area denotes the context
  that is available at decode time.}
  \label{fig:brnn}
\end{figure*}

We leverage the PixelRNN architecture \citep{Oord2016} and use a similar architecture (BinaryRNN) for the compression of binary codes of a single layer.
In this architecture (shown on \autoref{fig:brnn}), the estimation of the conditional code probabilities for line $y$ depends directly on some neighboring codes but also indirectly on the previously decoded binary codes through a line of states $S$ of size $1 \times W \times k$ which captures both some short term and long term dependencies. The state line is a summary of all the previous lines. In practice, we use $k = 64$. The probabilities are estimated and the state is updated line by line using a 1$\times$3 LSTM convolution.

The end-to-end probability estimation includes 3 stages.
First, the initial convolution is a 7$\times$7 convolution used to increase the receptive field of the LSTM state, the receptive field being the set of codes $c(i, j, \cdot)$ which can influence the probability estimation of codes $c(y, x, \cdot)$.
As in \citep{Oord2016}, this initial convolution is a masked convolution so as to avoid dependencies on future codes.
In the second stage, the line LSTM takes as input the result $z_0$ of this initial convolution and processes one scan line at a time.
Since LSTM hidden states are produced by processing the previous scan lines, the line LSTM captures both short- and long-term dependencies.
For the same reason, the input-to-state LSTM transform is also a masked convolution.
Finally, two 1$\times$1 convolutions are added to increase the capacity of the network to memorize more binary code patterns.
Since we attempt to predict binary codes, the Bernoulli-distribution parameter can be directly estimated using a sigmoid activation in the last convolution.

We want to minimize the number of bits used after entropy coding, which leads naturally to a cross-entropy loss.
In case of $\{0, 1\}$ binary codes, the cross-entropy loss can be written as:
\begin{equation}
  \sum_{y,x,d} -c \log_2(\hat{P}(c\mid T)) - (1-c) \log_2(1 - \hat{P}(c\mid T))
\end{equation}

\vspace{-0.5em}
\subsection{Progressive Entropy Coding}
\vspace{-0.5em}

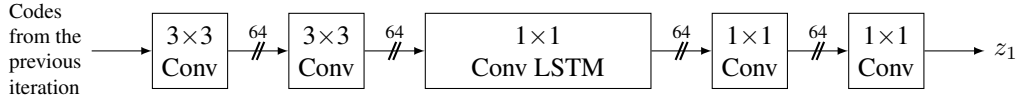
\begin{figure*}
  \centering
  \begin{tikzpicture}[scale=0.4,every text node part/.style={align=center},block/.style={draw,rectangle,minimum height=1.0cm,minimum width=1.0cm},strike thru arrow/.style={decoration={markings, mark=at position 0.5 with {\draw [thick,-] (-0.02cm,-0.12cm) -- ( 0.08cm, 0.12cm); \draw [thick,-] (-0.08cm,-0.12cm) -- ( 0.02cm, 0.12cm);}}, postaction={decorate},}]
    \node (A1) at (0,0) [block] {3$\times$3\\Conv};
    \node (A2) at ([xshift=2cm]A1.east) [anchor=west,block] {3$\times$3\\Conv};
    \node (A3) at ([xshift=2cm]A2.east) [anchor=west,block,minimum width=3.0cm] {1$\times$1\\Conv LSTM};
    \node (A4) at ([xshift=2cm]A3.east) [anchor=west,block] {1$\times$1\\Conv};
    \node (A5) at ([xshift=2cm]A4.east) [anchor=west,block] {1$\times$1\\Conv};

    \draw [latex-] (A1.west) -- ++(-2cm,0) node[left=-0.2cm] {\footnotesize \begin{tabular}{l}Codes\\from the\\previous\\iteration\end{tabular}};
    \draw [-latex,strike thru arrow] (A1.east) -- node [midway,above=0.05cm] {\scriptsize 64} (A2.west);
    \draw [-latex,strike thru arrow] (A2.east) -- node [midway,above=0.05cm] {\scriptsize 64} (A3.west);
    \draw [-latex,strike thru arrow] (A3.east) -- node [midway,above=0.05cm] {\scriptsize 64} (A4.west);
    \draw [-latex,strike thru arrow] (A4.east) -- node [midway,above=0.05cm] {\scriptsize 64} (A5.west);
    \draw [-latex] (A5.east) -- ++(2.0cm,0) node [right] {$z_1$};
  \end{tikzpicture}
  \caption{Description of neural network used to compute additional line LSTM inputs for progressive entropy coder. This allows propagation of information from the previous iterations to the current.}
  \label{fig:progressive-brnn}
\end{figure*}

When dealing with multiple iterations, a baseline entropy coder would be to duplicate the single iteration entropy coder as many times as there are iterations, each iteration having its own line LSTM. However, such an architecture would not capture the redundancy between the iterations.
We can augment the data that is passed to the line LSTM of iteration \#k with some information coming from the previous layers: the line LSTM in \autoref{fig:brnn} receives not just $z_0$ like in the single iteration approach but also $z_1$ estimated from the previous iteration using a recurrent network as shown on \autoref{fig:progressive-brnn}. Computing $z_1$ does not require any masked convolution since the codes of the previous layers are fully available.

\vspace{-0.5em}
\section{Results}
\vspace{-0.5em}
\label{section:results}

\textbf{Training Setup:} In order to evaluate the recurrent models we described, we used two sets of training data.
The first dataset is the \enquote{32$\times$32} dataset gathered in \citet{toderici2015variable}.
The second dataset takes a random sample of 6 million 1280$\times$720 images on the web, decomposes the images into non-overlapping 32$\times$32 tiles and samples 100 tiles that have the worst compression ratio when using the PNG compression algorithm.
By selecting the patches that compress the least under PNG, we intend to create a dataset with \enquote{hard-to-compress} data.
The hypothesis is that training on such patches should yield a better compression model.
We refer to this dataset as the \enquote{High Entropy (HE)} dataset.

All network architectures were trained using the Tensorflow~\citep{tensorflow} API, with the Adam~\citep{kingma:2014} optimizer. Each network was trained using learning rates of $\left[0.1, ... ,2\right]$. The $L_1$ loss (see \autoref{eq:loss}) was weighted by $\beta = \left(s\times n\right)^{-1}$ where $s$ is equal to $B\times H\times W\times C$ where $B=32$ is the batch size, $H=32$ and $W=32$ are the image height and width, and $C=3$ is the number of color channels. $n=16$ is the number of RNN unroll iterations.
%We employed a batch size of 32.

\textbf{Evaluation Metrics:} In order to assess the performance of our models, we use a
perceptual, full-reference image metric for comparing original, uncompressed
images to compressed, degraded ones. It is important to note that there is no
consensus in the field for which metric best represents human perception so
the best we can do is sample from the available choices while acknowledging
that each metric has its own strengths and weaknesses.
We use Multi-Scale Structural Similarity (MS-SSIM) \citep{wang2003multiscale}, a well-established
metric for comparing lossy image compression algorithms, and the more recent
Peak Signal to Noise Ratio - Human Visual System (PSNR-HVS) \citep{psnrhvs}.
We apply MS-SSIM to each of the RGB channels independently and average the results,
while PSNR-HVS already incorporates color information.
MS-SSIM gives a score between 0 and 1, and PSNR-HVS is measured in decibels.
In both cases, higher values imply a closer match between the test and reference images.
Both metrics are computed for all models over the reconstructed images after each iteration.
In order to rank models, we use an aggregate measure computed as the area under the rate-distortion curve (AUC).

We collect these metrics on the widely used Kodak Photo CD dataset~\citep{kodak}.
The dataset consists of 24 768$\times$512 PNG images
(landscape/portrait) which were never compressed with a lossy algorithm.

\textbf{Architectures:} We ran experiments consisting of \{GRU, Residual GRU,
LSTM, Associative LSTM\} $\times$ \{One Shot Reconstruction, Additive
Reconstruction, Additive Rescaled Residual\} and report the results for the
best performing models after 1 million training steps.

\begin{figure*}
  \centering
  \includegraphics[width=0.45\textwidth]{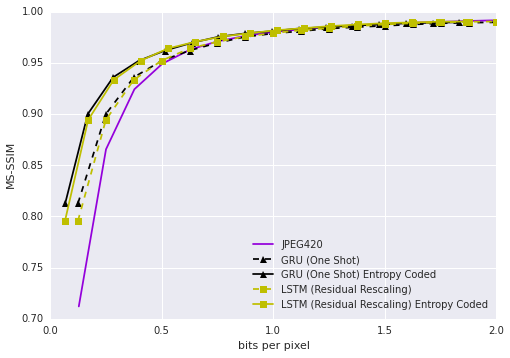}
  \hfill
  \includegraphics[width=0.45\textwidth]{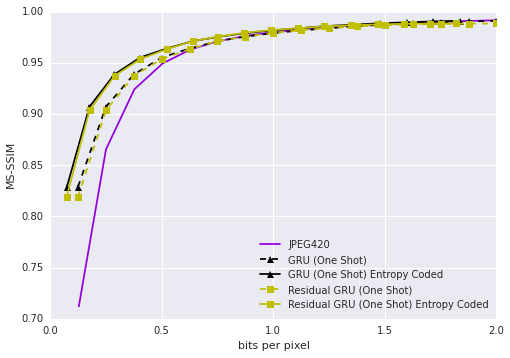}
  \caption{Rate distortion curve on the Kodak dataset given as MS-SSIM vs. bit per pixel (bpp). Dotted lines: before entropy coding, Plain lines: after entropy coding.
           Left: Two top performing models trained on the 32x32 dataset. Right: Two top performing models trained on the High Entropy dataset.}
  \label{fig:results-post-ec-ssim}
\end{figure*}

\begin{figure}
  \centering
  \includegraphics[width=0.45\textwidth]{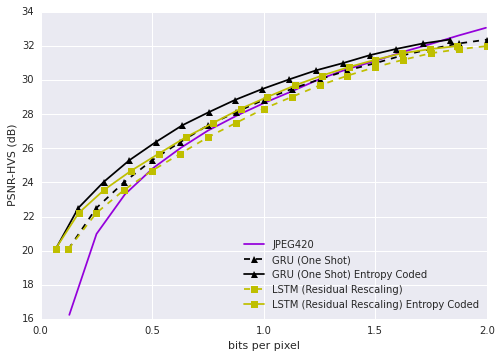}
  \hfill
  \includegraphics[width=0.45\textwidth]{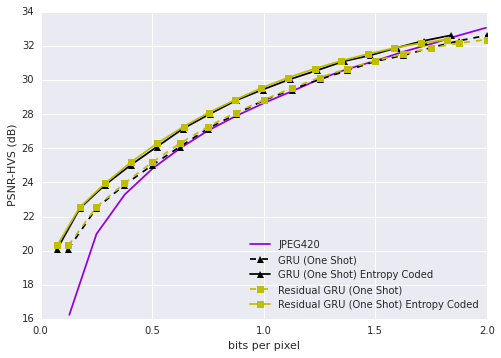}
  \caption{Rate distortion curve on the Kodak dataset given as PSNR-HVS vs. bit per pixel (bpp). Dotted lines: before entropy coding, Plain lines: after entropy coding.
           Top: Two top performing models trained on the 32x32 dataset. Bottom: Two top performing models trained on the High Entropy dataset.}
  \label{fig:results-post-ec-psnr-hvs}
\end{figure}

It is difficult to pick a \enquote{winning} architecture since the two metrics
that we are using don't always agree.
To further complicate matters, some models may perform better at low bit rates,
while others do better at high bit rates.
In order to be as fair as possible, we picked those models which had the largest area under the curve, and plotted them in
\autoref{fig:results-post-ec-ssim} and \autoref{fig:results-post-ec-psnr-hvs}.

The effect of the High Entropy training set can be seen in \autoref{table:auc}.
In general models benefited from being trained on this dataset rather than on
the 32$\times$32 dataset, suggesting that it is important to train models using
\enquote{hard} examples. For examples of compressed images from each method, we
refer the reader to the supplemental materials.

When using the 32$\times$32 training data, GRU (One Shot) had the highest
performance in both metrics. The LSTM model with Residual Scaling had the
second highest MS-SSIM, while the Residual GRU had the second highest PSNR-HVS.
When training on the High Entropy dataset, The One Shot version of LSTM had the
highest MS-SSIM, but the worst PSNR-HVS. The GRU with \enquote{one shot} reconstruction
ranked 2nd highest in both metrics, while the Residual GRU with \enquote{one shot}
reconstruction had the highest PSNR-HVS.

\begin{figure*}
  \centering
  \includegraphics[width=1.0\textwidth]{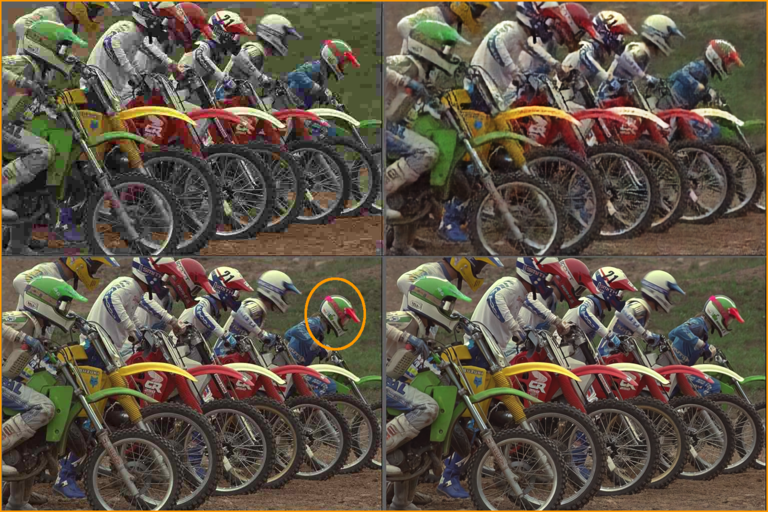}
  \caption{Comparison of compression results on Kodak Image 5. The top row is target at 0.25 bpp, the bottom row at 1.00 bpp. The left column is JPEG 420 and the right column is our Residual GRU (One Shot) method. The bitrates for our method are before entropy coding. In the first row (0.25 bpp) our results are more able to capture color (notice color blocking on JPEG). In the second row (1.00 bpp) our results don't incur the mosquito noise around objects (one example is highlighed with an orange circle). Results at 1 bpp may be difficult to see on printed page. Additional results available in the supplemental materials.}
  \label{fig:results-images}
\end{figure*}

We depict the results of compressing image 5 from the Kodak dataset
in \autoref{fig:results-images}. We invite the reader to refer to the
supplemental materials for more examples of compressed images from the Kodak
dataset.

\begin{table*}
  \centering
  \caption{Performance on the Kodak dataset measured as area under the curve (AUC) for the specified metric, up to 2 bits per pixel. All models are trained up for approximately 1,000,000 training steps. No entropy coding was used. After entropy coding, the AUC will be higher for the network-based approaches.}
  \label{table:auc}
\begin{tabular}{lrrrr}
  \toprule
  \multicolumn{5}{c}{Trained on the 32$\times$32 dataset.}\\
  \midrule
  Model & Rank &  MS-SSIM AUC &  Rank &  PSNR-HVS AUC \\
  \midrule
  \textbf{GRU (One Shot)}                & \textbf{1} &               \textbf{1.8098} & \textbf{1}&                \textbf{53.15} \\
  \textbf{LSTM (Residual Scaling)}     & \textbf{2} &               \textbf{1.8091} &          4 &                         52.36 \\
  LSTM (One Shot)                        &          3 &                        1.8062 &          3 &                         52.57 \\
  LSTM (Additive Reconstruction)         &          4 &                        1.8041 &          6 &                         52.22 \\
  \textbf{Residual GRU (One Shot)}       &          5 &                        1.8030 & \textbf{2} &                \textbf{52.73} \\
  Residual GRU (Residual Scaling)      &          6 &                        1.7983 &          8 &                         51.25 \\
  Associative LSTM (One Shot)            &          7 &                        1.7980 &          5 &                         52.33 \\
  GRU (Residual Scaling)               &          8 &                        1.7948 &          7 &                         51.37 \\

  \midrule
  Baseline \citep{toderici2015variable}  &            &                        1.7225 &            &                         48.36 \\

\midrule
\midrule
  \multicolumn{5}{c}{Trained on the High Entropy dataset.}\\
  \midrule
  \textbf{LSTM (One Shot)}                            &          \textbf{1} &                        \textbf{1.8166} &          8 &                         48.86 \\
  \textbf{GRU (One Shot)}                             &          \textbf{2} &                        \textbf{1.8139} &          \textbf{2} &                         \textbf{53.07} \\
  \textbf{Residual GRU (One Shot)}                    &          3 &                        1.8119 &          \textbf{1} &                         \textbf{53.19} \\
  Residual GRU (Residual Scaling)                   &          4 &                        1.8076 &          7 &                         49.61 \\
  LSTM (Residual Scaling)                           &          5 &                        1.8000 &          4 &                         51.25 \\
  LSTM (Additive)                                     &          6 &                        1.7953 &          5 &                         50.67 \\
  Associative LSTM (One Shot)                         &          7 &                        1.7912 &          3 &                         52.09 \\
  GRU (Residual Scaling)                              &          8 &                        1.8065 &          6 &                         49.97 \\

  \midrule
  Baseline LSTM \citep{toderici2015variable}          &            &                        1.7408 &            &                         48.88 \\
\midrule
\midrule
  \multicolumn{5}{c}{JPEG}\\
  \midrule
  YCbCr 4:4:4 & & 1.7748 & & 51.28 \\
  YCbCr 4:2:0 & & 1.7998 & & 52.61 \\
  \bottomrule
\end{tabular}
\end{table*}

\textbf{Entropy Coding:}
The progressive entropy coder is trained for a specific image encoder, and we compare a subset of our models.
For training, we use a set of 1280$\times$720 images that are encoded using one of the previous image encoders (resulting in a 80$\times$45$\times$32 bitmap or $\nicefrac18$ bits per pixel per RNN iteration).

\autoref{fig:results-post-ec-ssim} and \autoref{fig:results-post-ec-psnr-hvs} show that all models benefit from this additional entropy coding layer.
Since the Kodak dataset has relatively low resolution images, the gains are not very significant -- for the best models we gained between 5\% at 2 bpp, and 32\% at 0.25 bpp.
The benefit of such a model is truly realized only on large images.
We apply the entropy coding model to the Baseline LSTM model, and the bit-rate saving ranges from 25\% at 2 bpp to 57\% at 0.25 bpp.
% The better the patch-based encoder model is, the worse performance the BinaryRNN will have.

\vspace{-0.5em}
\section{Discussion}
\vspace{-0.5em}

We presented a general architecture for compressing with RNNs,
content-based residual scaling,
and a new variation of GRU, which provided the highest PSNR-HVS out of the models trained on the high entropy dataset.
Because our class of networks produce image distortions that are not well captured by the existing perceptual metrics, it is difficult to declare a best model.
However, we provided a set of models which perform well according to these metrics,
and on average we achieve better than JPEG performance on both MS-SSIM AUC and
PSNR-HVS AUC, both with and without entropy coding. With that said, our models do benefit from the additional
step of entropy coding due to the fact that in the early iterations the recurrent encoder models produce
spatially correlated codes. Additionally, we are open sourcing our best Residual GRU model and our Entropy Coder training and evaluation in \href{https://github.com/tensorflow/models/tree/master/compression}{https://github.com/tensorflow/models/tree/master/comp\
ression}.

The next challenge will be besting compression methods derived from video compression codecs,
such as WebP (which was derived from VP8 video codec), on large images since they employ tricks such as reusing patches that were already decoded. Additionally training the entropy coder (BinaryRNN) and the patch-based encoder jointly
and on larger patches should allow us to choose a trade-off between the
efficiency of the patch-based encoder and the predictive power of the entropy coder. Lastly, it is important to emphasize that the domain of perceptual differences
is in active development. None of the available perceptual metrics
truly correlate with human vision very well, and if they do, they only correlate
for particular types of distortions. If one such metric were capable of correlating
with human raters for all types of distortions, we could incorporate it directly
into our loss function, and optimize directly for it.
% Remove if needed: just add a page break since it didn't make sense to have only 2 lines of references
% before switching page. Introduce a page break in order to have all the references on the same page.
\newline
\newline
\vspace{-0.5em}
\section{Supplementary Materials}
Supplementary Materials are available here: \url{https://storage.googleapis.com/compression-ml/residual_gru_results/supplemental.pdf}.
%\href{https://storage.googleapis.com/compression%2dml/residual_gru_results/supplemental.pdf}{https://storage.googleapis.com/compression-ml/residual_gru_results/supplemental.pdf}.
\vspace{-0.5em}
\newline
\newline
\newline
\newline

{\small
\bibliographystyle{ieee}
\bibliography{biblio}
}

\end{document}

\appendix
\section{Supplementary Materials}
We include a sample of images from the Kodak dataset, to compare JPEG with the compression of our network at various bitrates.
On the first page of each subsection, we show the full original image followed (at the bottom of that page) by a detailed crop area.  The cropped areas are shown for the different approaches under consideration and for multiple bitrates.  The approaches shown, to the right of the original cropped image, are: JPEG 420, LSTM Residual Scaling (trained on the 32x32 dataset), GRU One-shot (trained on the 32x32 dataset), GRU One-shot (trained on the high-entropy dataset), and Residual GRU One-shot (trained on the 32x32 dataset).  The (nominal) bitrates shown are: 0.5 bits per pixel (bpp), 1.0 bpp, and 2.0 bpp.
\\
\\
The remaining pages of each subsection show the complete images, compressed using these different approaches at those same bitrates.
This comparison will be biased towards JPEG, since we do use nominal bitrates on our results and we do not include the header size in the JPEG evaluation.  Our nominal bitrates do not include entropy coding, so our final bitrates will actually be lower than those quoted here.  Based on experiments, the final bitrate will be about about 15\% lower than the nominal bitrate at 0.5 bpp and about 5\% lower at 2 bpp.  We also do not count the JPEG header size in our search for the best JPEG setting for these bitrates, so the final JPEG bitrates will be slightly higher than what is quoted here.
\clearpage

\subsection{Kodak Image 1}
\begin{tikzpicture}
    \node[anchor=south west,inner sep=0] (image) at (0,0) {\includegraphics[width=\textwidth]{images/kodak/reference/kodim01.png}};
    \begin{scope}[x={(image.south east)},y={(image.north west)}]
        \draw[green,line width=0.25mm] (0.3242,0.455) rectangle (0.3659,0.5176);
    \end{scope}
\end{tikzpicture}
\newline
\newcommand\includekodak[1]{\includegraphics[trim=249 233 487 247,clip=true,width=\zoomimagewidth]{images/kodak/#1}}
\includekodakpatches{kodim01}

\subsection{Kodak Image 8}
\begin{tikzpicture}
    \node[anchor=south west,inner sep=0] (image) at (0,0) {\includegraphics[width=\textwidth]{images/kodak/reference/kodim08.png}};
    \begin{scope}[x={(image.south east)},y={(image.north west)}]
        \draw[red,line width=0.25mm] (0.7070,0.3027) rectangle (0.7487,0.3652);
    \end{scope}
\end{tikzpicture}
\newline
\renewcommand\includekodak[1]{\includegraphics[trim=543 155 193 325,clip=true,width=\zoomimagewidth]{images/kodak/#1}}
\includekodakpatches{kodim08}

\subsection{Kodak Image 24}
\begin{tikzpicture}
    \node[anchor=south west,inner sep=0] (image) at (0,0) {\includegraphics[width=\textwidth]{images/kodak/reference/kodim24.png}};
    \begin{scope}[x={(image.south east)},y={(image.north west)}]
        \draw[red,line width=0.25mm] (0.8099,0.0898) rectangle (0.8516,0.1523);
    \end{scope}
\end{tikzpicture}
\newline
\renewcommand\includekodak[1]{\includegraphics[trim=622 46 114 434,clip=true,width=\zoomimagewidth]{images/kodak/#1}}
\includekodakpatches{kodim24}